\documentclass{article}
     \usepackage[preprint, nonatbib]{tackling_climate_workshop_style}
\usepackage[utf8]{inputenc} 
\usepackage[T1]{fontenc}    
\usepackage{hyperref}       
\usepackage{url}            
\usepackage{booktabs}       
\usepackage{amsfonts}       
\usepackage{nicefrac}       
\usepackage{microtype}      
\usepackage{graphicx, subfig}
\usepackage[export]{adjustbox}
\usepackage{pifont}
\usepackage{xcolor}
\definecolor{mpgorange}{RGB}{239,124,0}
\definecolor{mpggreendark}{RGB}{0,85,85}
\title{Comparing Data-Driven and Mechanistic Models for Predicting 
    Phenology in Deciduous Broadleaf Forests}

\author{%
    Christian Reimers, David Hafezi Rachti, Guahua Liu and Alexander J. 
    Winkler\\
    Max Planck Institute for Biogeochemistry\\
    Hans-Kn\"oll-Str. 10\\
    {\tt\small \{creimers, drachti, gliu, awinkler\}@bgc-jena.mpg.de}
}

\begin{document}

\twocolumn[{%
      \maketitle
\begin{abstract}
    Understanding the future climate is crucial for informed policy decisions on 
    climate change prevention and mitigation. Earth system models play an 
    important role in predicting future climate, requiring accurate 
    representation of complex sub-processes that span multiple time scales and 
    spatial scales. One such process that links seasonal and interannual climate 
    variability to cyclical biological events is tree phenology in deciduous 
    broadleaf forests. Phenological dates, such as the start and end of the 
    growing season, are critical for understanding the exchange of carbon and 
    water between the biosphere and the atmosphere. Mechanistic prediction of 
    these dates is challenging. Hybrid modelling, which integrates data-driven 
    approaches into complex models, offers a solution. In this work, as a first 
    step towards this goal, train a deep neural network to predict a 
    phenological index from meteorological time series. We find that this 
    approach outperforms traditional process-based models. This 
    highlights the potential of data-driven methods to improve climate 
    predictions. We also analyze which variables and aspects of the time series 
    influence the predicted onset of the season, in order to gain a better 
    understanding of the advantages and limitations of our model.
\end{abstract}
    \vspace{2em}
}]

\section{Introduction}
\begin{figure*}[htb]
    \includegraphics[width=\textwidth]{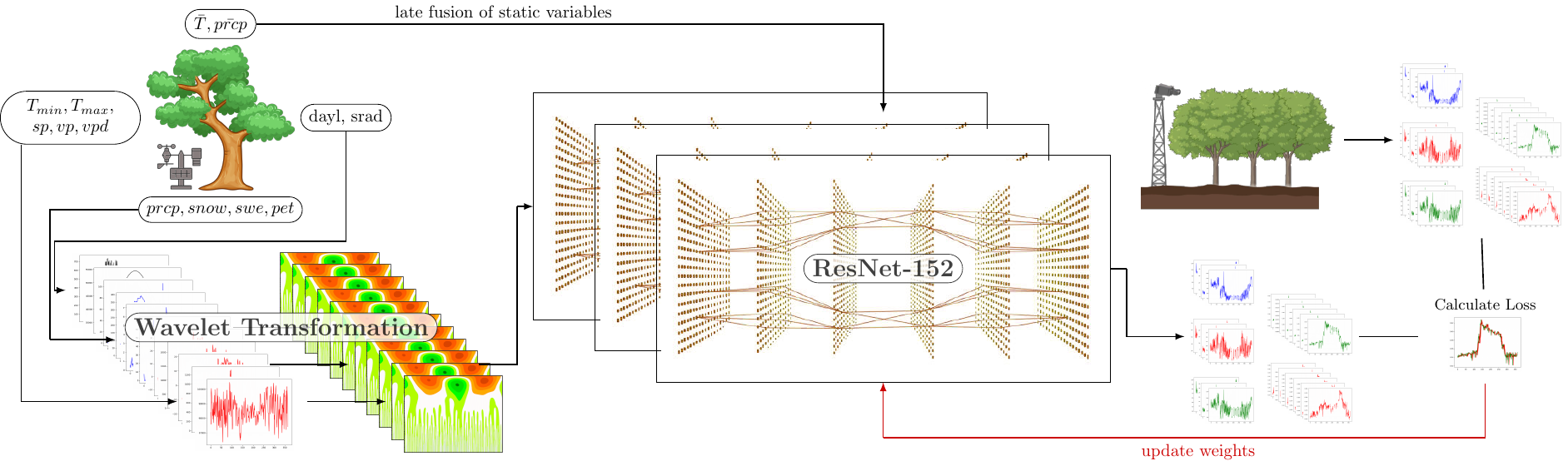}
    \caption{
        Our approach to predicting the green chromatic coordinate from 
        meteorological variables. We use a wavelet transform on 
        meteorological data from the current and previous years and an 
        ensemble of ResNets to predict phenology and several auxiliary 
        labels.}
    \label{fig:teaser}
\end{figure*}


    Understanding the future climate is important towards making political 
    decisions for climate change prevention and mitigation. The main method to 
    predict future climate are large Earth’s system models (ESMs). These models 
    need to represent many complex sub-processes that span  
    multiple time scales and spacial scales to make good predictions. One such 
    process is tree phenology in deciduous broadleaf forest. Phenology describes 
    the timing of periodic events in biological life cycles and links them to 
    seasonal and interannual variations in climate\cite{phenology}, for example 
    the date when trees start to grow or shed their leafs. Since understanding 
    the length of the season in which the trees carry functional leafs is 
    essential to understand the exchange of carbon and water between the 
    biosphere and the atmosphere \cite{wingate2015interpreting}, in this work, 
    we aim to predict the stat of this season (SoS) and end of this season 
    (EoS).


    Two challenges emerge when predicting these phenological dates. First, the 
    change between dormancy and maximum leafs is not instantaneous but the 
    number and maturity of leaves increases gradually. Hence, researchers have 
    to rely on thresholds. However, previous work demonstrated that the 
    choice of threshold has a significant influence on interannual patterns and 
    long-term trends of these dates \cite{panwar2023methodological}. Second, the 
    processes that relate meteorological events to these dates are non-linear, 
    complex and not fully understood. 
    Nevertheless, ESMs often rely on simple models that use only very few simple 
    meteorological features, for example \cite{reick2021jsbach}, or even 
    prescribed phenology, for example \cite{baker2008seasonal, 
    schaefer2008combined, zhan2003analytical}.

%
    One approach that can aid in addressing these challenges is hybrid modeling, 
    as suggested by \cite{reichstein2019deep} and applied in similar situations, 
    for example, by \cite{elghawi2023hybrid, grundner2022deep}. Hybrid modeling 
    is an approach where, in a complex model, some parts are replaced by 
    data-driven models. For phenology, large datasets are available from 
    satellite observations \cite{wang2023estimation}, in-situ studies 
    \cite{liu2021higher}, and near-remote sensing \cite{wingate2015interpreting, 
    seyednasrollah2019phenocam}. In this work, we present a data-driven model 
    for phenology. This is a first step towards replacing the phenology model in 
    a land surface model (LSM). We train a convolutional neural network (CNN), 
    namely a ResNet-152\cite{he2016deep}, on near-remote sensing observations of 
    a plant phenological index in deciduous broadleaf forest.
    
    Near-remote sensing data is available in higher 
    temporal resolutions and is more objective than in-situ observations
    \cite{liu2021higher}. Further, they are less susceptible to atmospheric 
    disturbance like clouds than satellite observations 
    \cite{richardson2007use}. 
    To tackle the challenges mentioned above, we, first, not only predict the 
    phenological dates SoS and EoS, but daily greenness indices, allowing us to 
    calculate the dates with different thresholds post-hoc. Second, we use a 
    ResNet, which can learn complex relations between meteorological time 
    series and the phenological index from the data.


    We compare our data-driven approach to two process-based approaches. First, 
    LoGrop-P, the phenology model of JSBACH \cite{reick2021jsbach}, the LSM of 
    ICON-ESM \cite{jungclaus2022icon}; and second, a model that prescribes 
    phenology. We find that the data-driven model reduces the error compared to 
    these models for daily greenness by 16\% and for the start of season (SoS) 
    by 47\% / 9\%. However, we find no improvement in predicting the end of 
    season (EoS) and, in an ablation study, we find that our approach is only 
    slightly better than simpler data-driven models.

   
    Finally, we use two methods to interpret the neural network. First, we use 
    Integrated Gradients (IG) \cite{sundararajan2017axiomatic} to understand 
    which variables have the largest influence towards the start of season. To 
    understand, in particular, which time scales are important, we perform a 
    wavelet transformation (WT) on the input time series before using the CNN. 
    We find that the network relies primarily on features and scales that are 
    more informative of the general climate than meteorological events. 
    
    Secondly, we employ an approach based on causal inference
    \cite{reimers2020determining} to validate that the CNN relies on 
    growing degree days (GDD) and chilling days two higher level descriptors of 
    the temperature time-series that are known to be important towards the SoS 
    and are used by the LoGro-P model.

\section{Related Work}
    

    Predicting the greenness of forests using machine learning has been 
    attempted before. For example, \cite{wheeler2023predicting} conducted a 
    challenge where teams predicted the greenness of multiple different sites up
    to 30 days ahead. In contrast to this work, where we train one global model,
    the teams in the challenge trained one model per site. Further, they did
    not use deep learning, but mostly simpler data-driven approaches.
    In contrast, \cite{guan2021analysis} trained a Long Short Term Memory model 
    but also only on a single site in the US.


\section{Data and Methods}
    \label{sec:met}\label{sec:model}
        
%
%

    To calculate the phenological state of the forest, we use the green 
    chromatic coordinate (GCC) \cite{richardson2007use} from observations 
    provided by the PhenoCam network \cite{seyednasrollah2019phenocam}. 
    Digital images are captured at various locations throughout North America at 
    intervals ranging from half-hourly to daily.
    For each site, a region of interest containing the canopy is selected, and 
    the mean RGB color is calculated. The GCC is given by 
    \begin{equation}
        GCC = G / (R + G + B).
    \end{equation}
    Following previous work \cite{wingate2015interpreting}, we use the 90$^{th}$ 
    percentile of GCC for each day as the target. The GCC dataset we use is 
    provided by the PhenoCam network \cite{seyednasrollah2019phenocam}. We use 
    the data from all deciduous broadleaf forest sites in the network that, 
    contain at least one year of observations,  are in the top two data-quality 
    groups and contain no GCC value below 0.1 or above 
    0.6. The data is standardized such that the smallest value becomes zero and 
    the largest value becomes one. We split the sites randomly into 66 sites for 
    training (247 site years), 9 sites for validation (37 site years), and 16 
    sites for testing (73 site years). 

    As input, we use eleven meteorological 
    variables: day length, precipitation, short-wave radiation, snow-water 
    equivalent, vapor pressure, vapor pressure deficit (VPD), minimum and 
    maximum air temperature (T$_{min}$, T$_{max}$), potential evapotranspiration, snow, 
    and surface air pressure. The first 
    ten variables are part of Daymet project\cite{thornton2022daymet}, a project
    that interpolates meteorological variables from ground-based measurements,
    and the air pressure is taken from ERA-5 \cite{hersbach2020era5} reanalysis 
    data.



    Following \cite{zeng2023transformers}, we decided for a direct rather than 
    iterative multi-step prediction. We predict the 365 
    yearly values of GCC simultaneously using one neural network with 365 
    outputs. Further, to account for legacy effects in vegetation, for example, 
    reported in \cite{yu2022contrasting}, we use the meteorological observations 
    of the current and the previous year, leaving us with eleven time series of 
    730 days. 
    We use a continuous wavelet transformation on each of the time 
    series individually. We use the Ricker-wavelet \cite{ricker1943further} with 
    a scale of $2^i$ days for $i \in \{-1, \dots, 8\}$ positioned at each day of 
    the time series using zero padding.
    
    We train a ResNet \cite{he2016deep} on the wavelet-transformed time series.
    We stack the results for the ten scales for all eleven variables along the
    first axis to create a 108 by 730 input with one channel. All input values 
    are normalized to mean zero and standard deviation one.
    Further, we use a late fusion of the annual temperature and precipitation
    means over 30 years after the convolutional part of the ResNet.
    We pre-train on ImageNet \cite{russakovsky2015imagenet} and afterwards alter 
    the CNN to the suitable input and output dimensions. 

    Using \cite{bergstra2011algorithms}, we optimize the size of the ResNet 
    (\{18, 34, 50, 101, 152\}; optimal: 152) along other hyper-parameters on the 
    validation set. The other hyper-parameters are batch size ([1, 128];
    optimal: 1), learning rate ([0.00001, 1]; optimal: 0.912), the parameters of 
    the cosine annealing with warm restarts ($t_0$: [10, 1000]; optimal: 790), 
    the optimizer used (\{Adam, SGD\}; optimal: SGD), and the number of epochs 
    before training stops early ([1, 100]; optimal: 18).

%

    To avoid over-fitting, we use multiple auxiliary tasks. In addition to the
    90$^{th}$ GCC percentile, we predict 20 additional color indices provided
    by the PhenoCam network, as well as five yearly values for kernel-NDVI
    \cite{wang2023estimation} from MODIS \cite{modis}: mean, standard deviation, 
    and 50$^{th}$, 75$^{th}$, and 90$^{th}$ percentile. We standardize all 
    labels and minimize the sum over the individual mean squared errors. To 
    focus on GCC, we multiply the loss of the auxiliary tasks with a constant $0 
    < \lambda < 1$, which we optimized as an additional hyper-parameter on the 
    validation set ([0,1]; optimal: 0.890). An overview is displayed in 
    Figure~\ref{fig:teaser}. 
    Finally, we add a random walk as an additional input. Later, we want to 
    interpret the importance of each variable, time-point, and scale using IG. 
    To this end, we dismiss any value that is less important than the 99.9 
    percentile of the random walk's importance scores.

    To reach a better generalization performance, we train twenty of these
    neural networks and use them in an ensemble for inference. To this end, we
    calculate the final prediction as a weighted mean over the predictions of 
    the individual models. The weights are optimized using the validation set.

    Additionally, we want to understand whether our data-driven model relies on
    high level features of the temperature time-series. Since these features are 
    not input variables, we cannot use IG or comparable methods to determine 
    whether they are relevant to the decision of the neural network. Instead we 
    use \cite{reimers2020determining}, a method based on causal inference.
    This method reduces the question whether a feature is used by a deep neural 
    network to a conditional dependence test. We use the 
    HSConIC \cite{fukumizu2007kernel} for this conditional dependence test with 
    a level of significance of $p = 0.05$. 
    Following \cite{reimers2021conditional}, we 
    validate that this method is suitable, by adding two features that do not 
    influence the SoS but are of similar complexity to the features of interest.

\section{Experiments and Results}
    

    To understand the quality and limitations of our data-driven model, we
    conduct four experiments: First, we evaluate our model against two 
    mechanistic models; second, we conduct an ablation study on the different 
    parts of our model; third, we quantify the importance of each variable 
    towards the SoS; and finally, we evaluate whether the network uses the same 
    features as mechanistic models.

    We compare the data-driven model
    (last line of Table~\ref{tab:numbers}) to two mechanistic models. The first 
    model uses a prescribed phenology (first line of Table~\ref{tab:numbers}). 
    For this model we prescribe $\mathbb E(GCC\,|\,DoY)$ the average GCC per day 
    of year on the training set. 
    The second model is LoGro-P, the phenology model 
    of JSBACH \cite{reick2021jsbach} (second line of Table~\ref{tab:numbers}). 
    
    For these models we compare the coefficient of 
    determination between predictions and observations for GCC (R$^2$), on the 
    GCC anomalies (R$^2_{\textup{anomalies}}$), on the start of season (SoS 
    R$^2$), and end of season
    (EoS R$^2$). Additionally, we report the root mean squared error for the GCC 
    values (RMSE), the SoS (SoS RMSE), and the EoS (EoS RMSE).

    To estimate the 
    start and end of season, we use the halfway point between the 
    highest and lowest GCC as a threshold. The LoGro-P model predicts the 
    leaf area index instead of GCC but it is highly related to GCC
    \cite{wingate2015interpreting}. 
    For this reason and because of the difficulties mentioned in 
    \cite{panwar2023methodological}, we calibrate the estimations of the SoS and 
    EoS of each model by training a linear correction on the validation set. 

    The results of the comparison are presented in Table~\ref{tab:numbers}. 
    The data-driven approach outperforms the mechanistic models for GCC and SoS,
    but not for EoS. The model increases R$^2$ by $13\%$ and reduces the RMSE by 
    $16\%$ compared to prescribing phenology. For the SoS, the model reduces the 
    RMSE by $47\%$ (7.0 days) compared to the prescribed phenology and by $9\%$ 
    (0.9 days) compared to LoGro-P.
    However, the coefficient of determination on the anomalies is only $0.198$,
    and, for EoS, the error increased by $3\%$ (1.2 days). 

    In the ablation study, we remove the wavelet transformation and/or replace 
    the ResNet by linear regression. We find that the simpler models have a 
    similar performance. The $R^2$ of the full model is identical to the linear
    regression and the SoS RSME of the linear regression is 3.7\% higher than 
    for the full model. Only R$^2_{\textup{anomalies}}$ improved notably 
    ($27\%$).

    \begin{table*}
        \centering
        \caption{The performance on the twelve holdout test sets. The best value 
        in each column is \textbf{bold}} 
        \label{tab:numbers}
        \footnotesize
        \begin{tabular}{lccccccc}
            \toprule
            Model & R$^2$ & R$^2_{\textup{anomalies}}$ & RMSE & 
                SoS R$^2$ & SoS RMSE & EoS R$^2$ & EoS RMSE \\
            \midrule
            Prescribed Phenology & 0.599 & 0.000 & 0.031 & 
                0.000 & 14.9 & \textbf{0.000} & \textbf{34.9} \\ 
            LoGro-P\cite{reick2021jsbach} & -- & -- & -- & 
                0.688 & 8.7 & -0.010 & \textbf{34.9} \\
            Linear Regression & 0.661 & 0.155 & \textbf{0.026} & 
                0.697 & 8.2 & -0.056 & 35.2 \\
            \quad + Wavelet & 0.630 & 0.078 & 0.028 & 
                0.632 & 9.0 & -0.024 & 35.8 \\
            ResNet & 0.643 & 0.112 & 0.028 & 
                0.716 & \textbf{7.9} & -0.094 & 37.0 \\
            \quad + Wavelet & \textbf{0.678} & \textbf{0.198} & \textbf{0.026} &  
                \textbf{0.720} & \textbf{7.9} & -0.025 & 36.1 \\
            \bottomrule
        \end{tabular}
    \end{table*}

\begin{figure}[tb!]
        \includegraphics[width=0.5\textwidth, height = 0.25\textwidth]{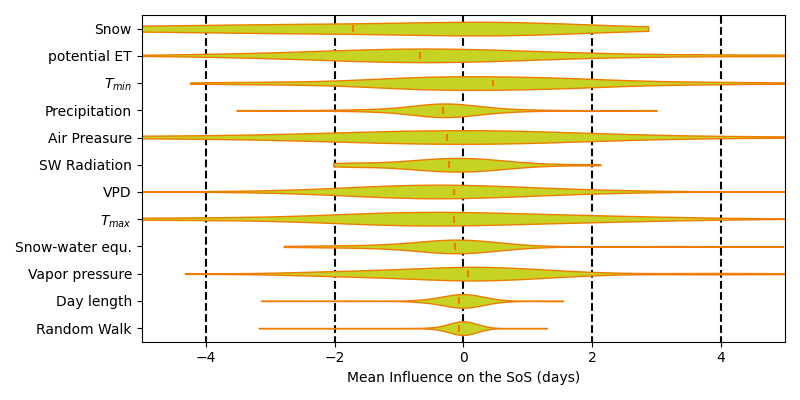}
        \caption{The importance of 
            each variable as evaluated by IG. The variables are ordered by 
            absolute mean importance. Each row shows the distribution of 
            influences on the SoS over all site-years. The orange line marks the 
            mean.
            }
        \label{fig:ig_results}
    \end{figure}

    Third, we want to quantify the importance of each variable towards the 
    prediction of SoS by IG \cite{sundararajan2017axiomatic}. However, the SoS 
    is not directly the output of the neural network. To be able to use IG, we 
    add an additional layer to the CNN,
    calculating 
    \begin{equation}
        150 - \sum_{t=80}^{150}\sigma(\textup{gcc}_t - 0.5)
    \end{equation}
    with $\sigma$ the sigmoid function. The result of this calculation is 
    virtually identical to the values determined by the threshold method 
    ($R^2 = 0.986$).
    For the importance analysis, we exclude years where the start of season is
    predicted before day 80. The results are displayed in 
    Figure~\ref{fig:ig_results}. The most important variables are the snow, the 
    potential evapotranspiration, which is a predictor of the soil moisture,  
    and the minimum temperature. While these features are known to be meaningful
    towards the SoS, we find that all days are of similar importance and long, 
    close to yearly, periods are much more relevant than daily periods, 
    indicating that the network relies on a general climate rather than specific 
    meteorological events.

    We compare this result, to the coefficients of the linear 
    classifier. Following \cite{jung2017compensatory} we multiply
    each linear coefficient with the standard deviation of the variable. We 
    consider the linear classifier for day of year (DoY) 120. Similar to our model, the linear
    classifier also relies on all days and not specifically on the DoYs directly 
    prior to DoY 120. Further, the three most importance features according to
    the linear classifier are the air pressure, the vapor pressure and the vapor 
    pressure deficit. The minimum temperature, a feature that many mechanistic
    models use exclusively, is ranked eighths.
 
%
   

    \begin{figure}[tb!]
        \includegraphics[width=0.5\textwidth, height = 0.25\textwidth]{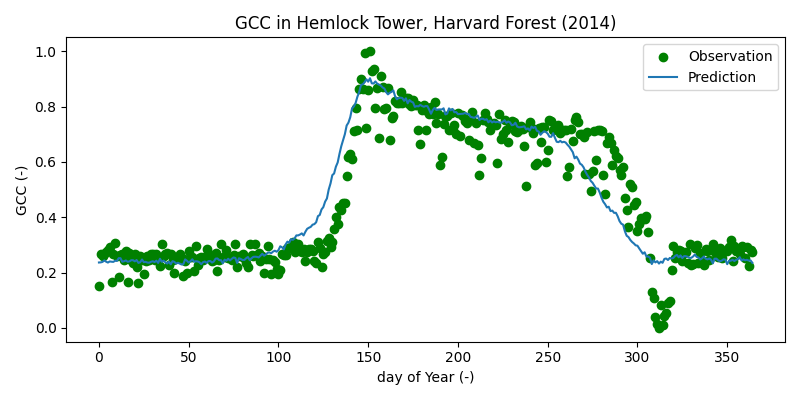} 
        \caption{The observation and prediction for one example from the test 
            set.}
        \label{fig:ig_results1}
    \end{figure}

    Finally we want to test whether our model relies on the same high-level 
    features as the mechanistic LoGro-P model. This model relies on two 
    features to predict the SoS \cite{reick2021jsbach}: the chill days, the 
    number of days below four degrees Celsius before a day of the year, and 
    the growing degree days  
    \begin{equation}
        GDD = \sum_{t = 1}^{doy} \max (0, T_t - 277.15 \textup{K})
    \end{equation}
    where $T_t$ is the mean temperature at day $t$. We determine 
    whether our model also relies on these features at DoY 120.
    To avoid detecting the importance of the average temperature, we decorrelate
    these features from from the average temperature. 
    Following \cite{reimers2021conditional}, we 
    validate that this method of \cite{reimers2020determining} is suitable, by 
    testing two additional features that do not 
    influence the SoS but are of similar complexity to the features of interest: 
    the chill days and GDD of the previous year.

    We find that the network uses the chill days and growing degree days of the 
    current but not the previous year to determine the SoS.

%

\section{Conclusions and Future Research}
    In this work we compare a data-driven method to two mechanistic models. 
    We find that our model performs better in predicting the greenness of 
    canopies and in predicting the start of season. 
    However, the architecture only slightly outperforms much simpler 
    architectures.
    We further analysed which features are used by the neural network to predict
    the start of season.
    While we found that our model considers the features used by the mechanistic
    LoGro-P model and considers variables relevant that are considered by 
    experts, the analysis with integrated gradients revealed that the model 
    considers mainly long time scales and does not focus on the time period 
    directly before the start of season.
    This indicates that the model understood the impact of the brought climate
    on the phenology but not the influence of meteorological events. 
    
    We found that no model can detect the end of season with any 
    accuracy. This might be due to problems in the data that might already add
    a high amount of uncertainty to the day of year when the end of season is 
    observed. Further, the data has considerable differences between sites, 
    likely due to differences in the data acquisition, for example,  different 
    orientations and angles of the cameras.

    Therefore, this approach could be improved by considering data from 
    different sources or putting a stronger focus on normalizing the GCC data 
    according to site specific properties.

\section*{Acknowledgments}
    Funding for this study was provided by the European Research Council (ERC) 
    Synergy Grant ``Understanding and Modelling the Earth System with Machine 
    Learning (USMILE)'' under the Horizon 2020 research and innovation programme 
    (Grant agreement No. 855187)

\bibliographystyle{plain}
\bibliography{literature}

\begin{thebibliography}{10}

\bibitem{baker2008seasonal}
IT~Baker, L~Prihodko, AS~Denning, Michael Goulden, S~Miller, and HR~Da~Rocha.
\newblock Seasonal drought stress in the amazon: Reconciling models and
  observations.
\newblock {\em Journal of Geophysical Research: Biogeosciences}, 113(G1), 2008.

\bibitem{bergstra2011algorithms}
James Bergstra, R{\'e}mi Bardenet, Yoshua Bengio, and Bal{\'a}zs K{\'e}gl.
\newblock Algorithms for hyper-parameter optimization.
\newblock {\em Advances in neural information processing systems}, 24, 2011.

\bibitem{elghawi2023hybrid}
RedaReda ElGhawi, Basil Kraft, Christian Reimers, Markus Reichstein, Marco
  K{\"o}rner, Pierre Gentine, and Alexander~J WinklerWinkler.
\newblock Hybrid modeling of evapotranspiration: inferring stomatal and
  aerodynamic resistances using combined physics-based and machine learning.
\newblock {\em Environmental Research Letters}, 18(3):034039, 2023.

\bibitem{fukumizu2007kernel}
Kenji Fukumizu, Arthur Gretton, Xiaohai Sun, and Bernhard Sch{\"o}lkopf.
\newblock Kernel measures of conditional dependence.
\newblock {\em Advances in neural information processing systems}, 20, 2007.

\bibitem{grundner2022deep}
Arthur Grundner, Tom Beucler, Pierre Gentine, Fernando Iglesias-Suarez, Marco~A
  Giorgetta, and Veronika Eyring.
\newblock Deep learning based cloud cover parameterization for icon.
\newblock {\em Journal of Advances in Modeling Earth Systems},
  14(12):e2021MS002959, 2022.

\bibitem{guan2021analysis}
Peng Guan, Yili Zheng, and Guannan Lei.
\newblock Analysis of canopy phenology in man-made forests using near-earth
  remote sensing.
\newblock {\em Plant Methods}, 17(1):1--15, 2021.

\bibitem{he2016deep}
Kaiming He, Xiangyu Zhang, Shaoqing Ren, and Jian Sun.
\newblock Deep residual learning for image recognition.
\newblock In {\em Proceedings of the IEEE conference on computer vision and
  pattern recognition}, pages 770--778, 2016.

\bibitem{hersbach2020era5}
Hans Hersbach, Bill Bell, Paul Berrisford, Shoji Hirahara, Andr{\'a}s
  Hor{\'a}nyi, Joaqu{\'\i}n Mu{\~n}oz-Sabater, Julien Nicolas, Carole Peubey,
  Raluca Radu, Dinand Schepers, et~al.
\newblock The era5 global reanalysis.
\newblock {\em Quarterly Journal of the Royal Meteorological Society},
  146(730):1999--2049, 2020.

\bibitem{jung2017compensatory}
Martin Jung, Markus Reichstein, Christopher~R Schwalm, Chris Huntingford,
  Stephen Sitch, Anders Ahlstr{\"o}m, Almut Arneth, Gustau Camps-Valls,
  Philippe Ciais, Pierre Friedlingstein, et~al.
\newblock Compensatory water effects link yearly global land co2 sink changes
  to temperature.
\newblock {\em Nature}, 541(7638):516--520, 2017.

\bibitem{jungclaus2022icon}
Johann~H Jungclaus, Stephan~J Lorenz, Hauke Schmidt, Victor Brovkin, Nils
  Br{\"u}ggemann, Fatemeh Chegini, Traute Cr{\"u}ger, Philipp De-Vrese,
  Veronika Gayler, Marco~A Giorgetta, et~al.
\newblock The icon earth system model version 1.0.
\newblock {\em Journal of Advances in Modeling Earth Systems},
  14(4):e2021MS002813, 2022.

\bibitem{liu2021higher}
Guohua Liu, Isabelle Chuine, R{\'e}my Den{\'e}ch{\`e}re, Fr{\'e}d{\'e}ric Jean,
  Eric Dufr{\^e}ne, Ga{\"e}lle Vincent, Daniel Berveiller, and Nicolas
  Delpierre.
\newblock Higher sample sizes and observer inter-calibration are needed for
  reliable scoring of leaf phenology in trees.
\newblock {\em Journal of Ecology}, 109(6):2461--2474, 2021.

\bibitem{phenology}
Merriam-Webster.
\newblock Phenology.

\bibitem{modis}
{ORNL DAAC}.
\newblock Modis and viirs land products global subsetting and visualization
  tool, 2018.

\bibitem{panwar2023methodological}
Annu Panwar, Mirco Migliavacca, Jacob~A Nelson, Jos{\'e} Cort{\'e}s, Ana
  Bastos, Matthias Forkel, and Alexander~J Winkler.
\newblock Methodological challenges and new perspectives of shifting vegetation
  phenology in eddy covariance data.
\newblock {\em Scientific Reports}, 13(1):13885, 2023.

\bibitem{reichstein2019deep}
Markus Reichstein, Gustau Camps-Valls, Bjorn Stevens, Martin Jung, Joachim
  Denzler, Nuno Carvalhais, and fnm Prabhat.
\newblock Deep learning and process understanding for data-driven earth system
  science.
\newblock {\em Nature}, 566(7743):195--204, 2019.

\bibitem{reick2021jsbach}
Christian~H Reick, Veronika Gayler, Daniel Goll, Stefan Hagemann, Marvin
  Heidkamp, Julia~EMS Nabel, Thomas Raddatz, Erich Roeckner, Reiner Schnur, and
  Stiig Wilkenskjeld.
\newblock Jsbach 3-the land component of the mpi earth system model:
  documentation of version 3.2.
\newblock 2021.

\bibitem{reimers2021conditional}
Christian Reimers, Niklas Penzel, Paul Bodesheim, Jakob Runge, and Joachim
  Denzler.
\newblock Conditional dependence tests reveal the usage of abcd rule features
  and bias variables in automatic skin lesion classification.
\newblock In {\em Proceedings of the IEEE/CVF Conference on Computer Vision and
  Pattern Recognition}, pages 1810--1819, 2021.

\bibitem{reimers2020determining}
Christian Reimers, Jakob Runge, and Joachim Denzler.
\newblock Determining the relevance of features for deep neural networks.
\newblock In {\em European Conference on Computer Vision}, pages 330--346.
  Springer, 2020.

\bibitem{richardson2007use}
Andrew~D Richardson, Julian~P Jenkins, Bobby~H Braswell, David~Y Hollinger,
  Scott~V Ollinger, and Marie-Louise Smith.
\newblock Use of digital webcam images to track spring green-up in a deciduous
  broadleaf forest.
\newblock {\em Oecologia}, 152:323--334, 2007.

\bibitem{ricker1943further}
Norman Ricker.
\newblock Further developments in the wavelet theory of seismogram structure.
\newblock {\em Bulletin of the Seismological Society of America},
  33(3):197--228, 1943.

\bibitem{russakovsky2015imagenet}
Olga Russakovsky, Jia Deng, Hao Su, Jonathan Krause, Sanjeev Satheesh, Sean Ma,
  Zhiheng Huang, Andrej Karpathy, Aditya Khosla, Michael Bernstein, et~al.
\newblock Imagenet large scale visual recognition challenge.
\newblock {\em International journal of computer vision}, 115:211--252, 2015.

\bibitem{schaefer2008combined}
Kevin Schaefer, G~James Collatz, Pieter Tans, A~Scott Denning, Ian Baker, Joe
  Berry, Lara Prihodko, Neil Suits, and Andrew Philpott.
\newblock Combined simple biosphere/carnegie-ames-stanford approach terrestrial
  carbon cycle model.
\newblock {\em Journal of Geophysical Research: Biogeosciences}, 113(G3), 2008.

\bibitem{seyednasrollah2019phenocam}
B.~Seyednasrollah, A.M. Young, K.~Hufkens, T.~Milliman, M.A. Friedl,
  S.~Frolking, M.~Richardson, A.D. a nd~Abraha, D.W. Allen, M.~Apple, M.A.
  Arain, J.~Baker, J.M. Baker, D.~Baldocchi, C.J. Bernacchi, J.~Bhat~tacharjee,
  P.~Blanken, D.D. Bosch, R.~Boughton, E.H. Boughton, R.F. Brown, D.M.
  Browning, S.P. Brunsell, N. an d~Burns, M.~Cavagna, H.~Chu, P.E. Clark, B.J.
  Conrad, E.~Cremonese, D.~Debinski, A.R. Desai, R.~Diaz-D~elgado, L.~Duchesne,
  A.L. Dunn, D.M. Eissenstat, T.~El-Madany, D.S.S. Ellum, S.M. Ernest,
  L.~Esposito, A. an d~Fenstermaker, L.B. Flanagan, B.~Forsythe, J.~Gallagher,
  D.~Gianelle, T.~Griffis, P.~Groffman, J.~Gu, L. an d~Guillemot, M.~Halpin,
  P.J. Hanson, D.~Hemming, A.A. Hove, E.R. Humphreys, A.~Jaimes-Hernandez, A.A.
  Jaradat, J.~Johnson, E.~Keel, V.R. Kelly, J.W. Kirchner, P.B. Kirchner,
  M.~Knapp, M.~Krassovski, O.~Langvall, G.~Lanthier, G.l. Maire, E.~Magliulo,
  T.A. Martin, B.~McNeil, G.A. Meyer, M.~Migliavacca, B.~P. Mohanty, C.E.
  Moore, R.~Mudd, J.W. Munger, Z.E. Murrell, Z.~Nesic, H.S. Neufeld, T.L.
  O'Halloran, A.C. Oechel, W. a nd~Oishi, W.W. Oswald, T.D. Perkins, M.L. Reba,
  B.~Rundquist, B.R. Runkle, E.S. Russell, A.~Sadler, E.J. a nd~Saha, N.Z.
  Saliendra, L.~Schmalbeck, M.D. Schwartz, R.L. Scott, E.M. Smith,
  O.~Sonnentag, P.~Stoy, S.~Strachan, K.~Suvocarev, J.E. Thom, R.Q. Thomas,
  A.K. Van~den berg, R.~Vargas, J.~Verfaillie, C.S. Vogel, J.J. Walker,
  N.~Webb, P.~Wetzel, S.~Weyers, A.V. Whipple, T.G. Whitham, G.~Wohlfahrt, J.D.
  Wood, S.~Wo~lf, J.~Yang, X.~Yang, G.~Yenni, Y.~Zhang, Q.~Zhang, and D.~Zona.
\newblock Phenocam dataset v2.0: Vegetation phenology from digital camera
  imagery, 2000-2018.
\newblock 2019.

\bibitem{sundararajan2017axiomatic}
Mukund Sundararajan, Ankur Taly, and Qiqi Yan.
\newblock Axiomatic attribution for deep networks.
\newblock In {\em International conference on machine learning}, pages
  3319--3328. PMLR, 2017.

\bibitem{thornton2022daymet}
MM~Thornton, R~Shrestha, Y~Wei, PE~Thornton, SC~Kao, BE~Wilson, BW~Mayer,
  Y~Wei, R~Devarakonda, RS~Vose, et~al.
\newblock Daymet: daily surface weather data on a 1-km grid for north america,
  version 4 r1.
\newblock {\em ORNL DAAC, Oak Ridge, Tennessee, USA. Single Pixel Extraction
  Tool| Daymet (ornl. gov)}, 2022.

\bibitem{wang2023estimation}
Qiang Wang, {\'A}lvaro Moreno-Mart{\'\i}nez, Jordi Mu{\~n}oz-Mar{\'\i}, Manuel
  Campos-Taberner, and Gustau Camps-Valls.
\newblock Estimation of vegetation traits with kernel ndvi.
\newblock {\em ISPRS Journal of Photogrammetry and Remote Sensing},
  195:408--417, 2023.

\bibitem{wheeler2023predicting}
Kathryn Wheeler, Michael Dietze, David LeBauer, Jody Peters, Andrew~D.
  Richardson, R.~Quinn Thomas, Kai Zhu, Uttam Bhat, Stephan Munch,
  Raphaela~Floreani Buzbee, Min Chen, Benjamin Goldstein, Jessica~S. Guo, Dalei
  Hao, Chris Jones, Mira Kelly-Fair, Haoran Liu, Charlotte Malmborg, Naresh
  Neupane, Debasmita Pal, Arun Ross, Vaughn Shirey, Yiluan Song, McKalee Steen,
  Eric~A. Vance, Whitney~M. Woelmer, Jacob Wynne, and Luke Zachmann.
\newblock Predicting spring phenology in deciduous broadleaf forests: An open
  community forecast challenge.
\newblock {\em {SSRN} Electronic Journal}, 2023.

\bibitem{wingate2015interpreting}
Lisa Wingate, J{\'e}r{\^o}me Og{\'e}e, Edoardo Cremonese, Gianluca Filippa,
  Toshie Mizunuma, Mirco Migliavacca, Christophe Moisy, Matthew Wilkinson,
  Christine Moureaux, Georg Wohlfahrt, et~al.
\newblock Interpreting canopy development and physiology using a european
  phenology camera network at flux sites.
\newblock {\em Biogeosciences}, 12(20):5995--6015, 2015.

\bibitem{yu2022contrasting}
Xin Yu, Ren{\'e} Orth, Markus Reichstein, Michael Bahn, Anne Klosterhalfen,
  Alexander Knohl, Franziska Koebsch, Mirco Migliavacca, Martina Mund, Jacob~A
  Nelson, et~al.
\newblock Contrasting drought legacy effects on gross primary productivity in a
  mixed versus pure beech forest.
\newblock {\em Biogeosciences}, 19(17):4315--4329, 2022.

\bibitem{zeng2023transformers}
Ailing Zeng, Muxi Chen, Lei Zhang, and Qiang Xu.
\newblock Are transformers effective for time series forecasting?
\newblock In {\em Proceedings of the AAAI conference on artificial
  intelligence}, volume~37, pages 11121--11128, 2023.

\bibitem{zhan2003analytical}
Xiwu Zhan, Yongkang Xue, and G~James Collatz.
\newblock An analytical approach for estimating co2 and heat fluxes over the
  amazonian region.
\newblock {\em Ecological Modelling}, 162(1-2):97--117, 2003.

\end{thebibliography}

\end{document}